\documentclass{article}

\usepackage{arxiv}

\usepackage[utf8]{inputenc} % allow utf-8 input
\usepackage[T1]{fontenc}    % use 8-bit T1 fonts
\usepackage[hidelinks]{hyperref}
\usepackage{url}            % simple URL typesetting
\usepackage{booktabs}       % professional-quality tables
\usepackage{amsfonts}       % blackboard math symbols
\usepackage{nicefrac}       % compact symbols for 1/2, etc.
\usepackage{microtype}      % microtypography
\usepackage{lipsum}

\title{Exploiting Unsupervised Pre-training and Automated Feature Engineering for Low-resource Hate Speech Detection in Polish}

\author{Renard Korzeniowski \And Rafał Rolczyński  \And Przemysław Sadownik \And Tomasz Korbak \And Marcin Możejko \AND
\textnormal{(Sigmoidal)}}

\begin{document}
\maketitle

\begin{abstract}
This paper presents our contribution to PolEval 2019 Task 6: Hate speech and bullying detection. We describe three parallel approaches that we followed: fine-tuning a pre-trained ULMFiT model to our classification task, fine-tuning a pre-trained BERT model to our classification task, and using the TPOT library to find the optimal pipeline. We present results achieved by these three tools and review their advantages and disadvantages in terms of user experience. Our team placed second in subtask 2 with a shallow model found by TPOT: a~logistic regression classifier with non-trivial feature engineering. 
\end{abstract}

% keywords can be removed

\section{Introduction}
This paper presents our contribution to PolEval 2019 Task 6: Hate speech and bullying detection\footnote{http://poleval.pl/tasks/task6}. In the following three sections we describe three parallel approaches that we followed: fine-tuning a pre-trained ULMFiT model to our classification task, fine-tuning a pre-trained BERT model to our classification task, and using a TPOT solution to find the optimal pipeline. They instantiate two important trends in modern machine learning: automated machine learning (autoML) and transfer learning. AutoML and transfer learning are of particular importance for industry practitioners who struggle with data scarcity and development time constraints. We describe results achieved by these three tools and them from a machine learning engineer point of view, highlighting advantages and disadvantages in terms of user experience.

Our team placed second with a pipeline consisting of count-vectorizing the documents, recursive feature elimination guided by an extra-tree classifier with Gini criterion and dual logistic regression with L2 regularization. Our official results were micro-average F1 score 87.10\% and macro-average F1 score 46.45\%.

%%%%%%%%%%%%%%%%%%%%%%%%%%%%%%%%%%%%%%%%%%%%%%%%%%%%%%%%%%%%%%%%%%%%%%%%%%%%%%%

\section{TPOT}
\subsection{Introduction}
Tree-based Pipeline Optimization Tool \cite{tpot1,tpot2} is an autoML solution, which uses
evolutionary algorithms to design tree-shaped machine learning pipelines based on  operators defined in the \texttt{scikit-learn} library \cite{scikit-learn}.
Before passing data to \texttt{TPOT}, we transformed the sentences using scikit-learn's \texttt{CountVectorizer}.
We ran the \texttt{TPOT} with sparse configuration.

\subsection{First Subtask: Binary Classification}
For the first task we left \texttt{TPOT} for about 9 hours. We discovered that the time processed is less important than what parameters were applied. We only tried accuracy for the fitness score. The results achieved are presented in Table \ref{table:tpot1}.

\begin{table}[ht!]
  \centering

\begin{tabular}{l c} 
\toprule
\textbf{Metric} & \textbf{Value} \\
\midrule
Precision & 30.65\%  \\
Recall & 56.72\% \\
Balanced F-score & 39.79\% \\
Accuracy & 77.00\% \\
\bottomrule
\end{tabular}
\caption{Metrics for \texttt{TPOT} in the first subtask}
\label{table:tpot1}
\end{table}

The best pipeline consisted of count-vectorizing the documents, recursive feature elimination guided by an extra-tree classifier with Gini criterion and dual logistic regression with l2 regularization. 

To improve balanced F-score, we manually lowered the decision threshold to 0.7\%.

\subsection{Second Subtask: Multiclass Classification}

In the second task, after about 17 minutes of computation on a multi-threaded machine, our solution achieved results presented in Table~\ref{table:tpot2}.

\begin{table}[ht!]
  \centering

\begin{tabular}{l c} 
\toprule
\textbf{Metric} & \textbf{Value} \\
\midrule
Micro-Average F-score & 87.10\%  \\
Macro-Average F-score &  46.45\% \\
\bottomrule
\end{tabular}
\caption{Metrics for \texttt{TPOT} in the second subtask}
\label{table:tpot2}
\end{table}

After the competition, we verified that it could achieve these and higher scores reliably. As a fitness 
function, we tried \texttt{f1\_micro, f1\_macro} and accuracy scores, which turned out to have a little difference in the
outcome.

So far we have noticed that the method consistently chose either SVM or logistic regression with an optional pre-processing step. After several hours, the best solution we managed to achieve consisted of count-vectorization of
the documents, selection of the best 6\% features, one-hot encoding and logistic regression with the L2 penalty and 0.05 regularization strength as the final classification model. The results are presented in Table~\ref{table:tpot3}.

\begin{table}[ht!]
  \centering

\begin{tabular}{l c} 
\toprule
\textbf{Metric} & \textbf{Value} \\
\midrule
Micro-Average F-score &  87.60\%  \\
Macro-Average F-score &  50.20\% \\
\bottomrule
\end{tabular}
\caption{Our post-contest experiments}

\label{table:tpot3}
\end{table}

\subsection{Transfer}
Training best pipeline found in the second task on binary labels resulted in  metrics displayed in Table~\ref{table:tpot4}.

\begin{table}[ht!]
  \centering

\begin{tabular}{l c} 
\toprule
\textbf{Metric} & \textbf{Value} \\
\midrule
Precision & 32.58\%  \\
Recall & 64.18\% \\
Balanced F-score & 43.22\% \\
Accuracy & 77.40\% \\
\bottomrule
\end{tabular}
\caption{The final model}
\label{table:tpot4}
\end{table}

\subsection{Summary}
\texttt{TPOT} seems to find solutions reasonably fast for this kind of tasks. We assume that since all other solutions performed comparably well in the second task, we could achieve the level of the irreducible error for it. We also observed that it might provide significant improvement when ensembled with other methods including neural-network based models described in further sections.

\subsection{Ease of Use}
Out of all the solutions we have tried, in our opinion, \texttt{TPOT} was the easiest one to use. However one should note that with default values for the evolutionary algorithm (with 100 generations, population size of 100 and maximum evaluation time of single individual of 5 minutes), the overall optimization process can be quite long.

%%%%%%%%%%%%%%%%%%%%%%%%%%%%%%%%%%%%%%%%%%%%%%%%%%%%%%%%%%%%%%%%%%%%%%%%%%%%%

\section{ULMFiT}

\subsection{Introduction}

The main  \texttt{ULMFiT} \cite{ULMFIT} schema consists of:
\begin{itemize}
    \item[\textbullet] training language model on a huge dataset,
    \item[\textbullet] fine-tuning language model to a smaller task-specific dataset,
    \item[\textbullet] using language model trained on classification data to improve its understanding of input text during classification.
\end{itemize}

This idea makes extensive usage of two major Machine Learning concepts, namely: transfer learning (which has proved successful in computer vision \cite{CV_transfer_learning}) and semi-supervised learning \cite{semi_supervised_learning}.

The aim of this section is to determine how fitting a pre-trained language model on small task-specific dataset influences the performance of the classifier build on top of the pre-trained model and using it as an encoder.

\subsection{Architecture}
The model architecture used for experiments was \texttt{AWD\_LSTM} \cite{AWD_LSTM}. This model architecture has a strong, built-in regularization in the form of smartly applied dropout. Because of that, we have found it as a good candidate for transfer learning base that involved fitting pre-trained model on a small dataset.

\subsection{Training Procedure}
In the following subsection, we describe the procedure used to preprocess the data and train our \texttt{ULMFiT} solution. Two datasets were needed to train language model: a huge unlabelled corpus to teach a Polish language model and a smaller one for classification fine-tuning. In our case, the smaller dataset was just the training set provided by the competition organizers and as the huge language corpus we have used the last year's \texttt{PolEval} \cite{poleval2018} dataset for language modeling task\footnote{http://2018.poleval.pl/task3/task3\_train.txt.gz}. 

Tokenization and general preprocessing of our corpora was performed using a popular NLP tool \texttt{SpaCy} \cite{spacy}.

We have used one fit cycle policy \cite{OneFitCycle1, OneFitCycle2, OneFitCycle3} to train both language model and classifier since it increased the test set accuracy and lowered the time of training.

\subsection{Results}
Metrics used in the competition depended on the task. Our \texttt{ULMFiT} solution was used only in the first task of binary classification, and we trained ``Without fine-tuning’’ solution. Second result ``With fine-tuning’’ in Table~\ref{table:1} below is the submission of winning team n-waves.

\begin{table}[ht!]
\centering
\begin{tabular}{l c c c c}
\toprule
\textbf{Fine-tuning} & \textbf{Precision} & \textbf{Recall} & \textbf{F1} & \textbf{Accuracy} \\
\midrule
With & 66.67 & 52.24 & 58.58 & 90.10  \\
Without & 52.71 & 50.75 & 51.71 & 87.30  \\
\bottomrule
\end{tabular}
\caption{Binary classification}
\label{table:1}
\end{table}

As we can see fine-tuning had a significant impact on the final performance of the model classifier.

\subsection{Accessibility}
We used \texttt{FastAI} implementation of \texttt{ULMFiT}. It was user-friendly and easy to use. The only problems we encountered were occasional unexpected errors connected to memory management when training a language model on a huge unlabeled corpus.

%%%%%%%%%%%%%%%%%%%%%%%%%%%%%%%%%%%%%%%%%%%%%%%%%%%%%%%%%%%%%%%%%%%%%%%%%%%%%%

\section{BERT}

One of the latest milestones in NLP development was the release of \texttt{BERT} (Bidirectional Encoder Representations from Transformers, \cite{bert}). \texttt{BERT} is an unsupervised, deeply bidirectional system for language model pre-training. It has become a state of the art method for many NLP tasks. The models, which were pre-trained in an unsupervised manner on massive plain text corpus, are available for download and free reuse. Thanks to that \texttt{BERT} might serve as a provider of bidirectional contextual words representations which can be easily reused in any language-processing pipeline.

\subsection{Binary Classifier}
The most straight-forward way to use BERT representations is to apply them for the classification task. In order to do that, we have fine-tuned the classifier with minimal changes applied to the BERT model architecture during the training phase (the process is performed in a manner similar to Semi-supervised Sequence Learning and \texttt{ULMFiT} fine-tuning process). In our experiments we have used the \texttt{BERT}-Base Multilingual Cased model as a classification base. This pre-trained model supports 104 languages (including Polish) and has over 110M parameters.\footnote{The model is case-sensitive and performs much better than the uncased one.} It consists of a trained Transformer Encoder stack with 12 layers (the size of hidden state is 768) and Multi-Head Attention (12-heads) \cite{attention_is_all, bert}. At the top of the base model, we have added the classification softmax layer.\footnote{Add dropout layer with ratio 0.1 helps to prevent high overfitting.} The hyper-parameters of the fine-tuning process are presented in Table~\ref{table:bert-train-process} (the selected model).

\begin{table}[ht!]
\centering
\begin{tabular}{c c c c c}
  \toprule
    \textbf{Batch size} & \textbf{Learning rate} & \textbf{Epochs} & \textbf{Warm-up} & \textbf{Max sequence}\\
    \midrule
    32 & $2\cdot 10^{-5}$ & 3 & 0.1 & 128 \\
    \bottomrule
\end{tabular}
\caption{Hyper-parameters used in the fine-tuning process}
\label{table:bert-train-process}
\end{table}

\subsection{Results}
The results of our best model are presented in Table~\ref{table:bert}. In our experiment we have used \texttt{BERT} \emph{mean-pooled output} of the last hidden layer which assigns a single vector to an entire sequence. In the future, we plan to experiment with several different pooling strategies (e.g. six choices examined by \cite{bert}).

\begin{table}[ht!]
\centering
\begin{tabular}{l c}
  \toprule
    \textbf{Metric} & \textbf{Value} \\
    \midrule
    Precision & 65.12\%  \\
    Recall & 42.31\% \\
    \textbf{Balanced F-score} & \textbf{51.32\%} \\
    Accuracy & 93.23\% \\
    \bottomrule
\end{tabular}
\caption{\texttt{BERT}: Binary Classification}
\label{table:bert}
\end{table}

\texttt{BERT} is a powerful component which can be used effectively in different types of tasks. Unfortunately, the largest version of the model, which currently is reported to achieve the state of the art results is ridiculously large (340M parameters) and it is unavailable for the multilingual use case. It is also worth to point out that it is currently impossible to reproduce most of the \texttt{BERT} results using \texttt{GPU} machine due to the out-of-memory issues what could be a severe limitation in everyday applications.

%%%%%%%%%%%%%%%%%%%%%%%%%%%%%%%%%%%%%%%%%%%%%%%%%%%%%%%%%%%%%%%%%%%%%%%%%%%%%

\section{Conclusions}

We arrived at a slightly surprising result that it was a shallow model produced by TPOT -- the easiest to use library we tried -- that earned us the second place in subtask two. Even if the reliability of these results can be questioned, it still proves a strong case in favor of proper exploitation of the powers of shallow models when training data are limited.  
\bibliographystyle{unsrt}  

\bibliography{poleval}

\end{document}